# Can Large Language Models Detect Verbal Indicators of Romantic Attraction?

Matz, S.C., Peters, H., Cerf, M. Grunenberg, E., Eastwick, P.W., Back, M.D., & Finkel, E.J.

As artificial intelligence (AI) models become an integral part of everyday life, our interactions with them shift from purely functional exchanges to more relational experiences. For these experiences to be successful, artificial agents need to be able to detect and interpret social cues and interpersonal dynamics - both within and outside of their own human-agent relationships. In this paper, we explore whether AI models can accurately decode one of the arguably most important but complex social signals: romantic attraction. Specifically, we test whether Large Language Models can detect romantic attraction during brief getting-to-know-you interactions between humans. Examining data from 964 speed dates, we show that ChatGPT can predict both objective and subjective indicators of speed dating success ($r$=0.12-0.23). Although predictive performance remains relatively low, ChatGPT's predictions of actual matching (i.e., the exchange of contact information) were not only on par with those of human judges but incremental to speed daters' own predictions. In addition, ChatGPT's judgments showed substantial overlap with those made by human observers ($r$=0.21-0.35), highlighting similarities in their representation of romantic attraction that are independent of accuracy. Our findings also offer insights into how ChatGPT arrives at its predictions and the mistakes it makes. Specifically, we use a Brunswik lens approach to identify the linguistic and conversational cues utilized by ChatGPT (and human judges) vis-a-vis those that are predictive of actual matching.

*Keywords:* Large Language Models, ChatGPT, speed-dating, romantic attraction, relational agents.

# Introduction

As artificial intelligence (AI) models have become an integral part of everyday life, our interactions with them are rapidly shifting from purely functional exchanges to more relational experiences. AI models – whether in the context of virtual assistants, chatbots, or social robots – are increasingly designed to mimic human-like abilities and characteristics and to act as conversation partners (Mariani et al., 2023). Consequently, we find ourselves interacting with them not just as tools but as social entities, engaging with them in ways that resemble how we interact with other humans.

A critical component of successful human-to-human interaction is the ability to detect and interpret social cues and interpersonal dynamics (Frith & Frith, 1999). Such interpersonal perceptions are a ubiquitous part of our everyday social lives (Back, 2021; Kenny, 2019).On a regular basis and without much cognitive effort, we interpret the appearances and behaviors of those around us – i.e., their tone of voice, facial expressions, or linguistic subtleties – in order to infer their personality, emotions and intentions (Funder, 1995; Nestler & Back, 2013).

Similar to human actors, AI models that take on relational roles need to be able to accurately interpret social signals. An AI taking on the role of a romantic partner, for example, needs to accurately decode the boredom or excitement of their human counterparts to make the conversation successful. An AI friend needs to understand the social dynamics of close relationships and larger social groups to participate in the latest gossip. And an AI leadership coach needs to be able to detect the frustration expressed by employees in a recorded business meeting to offer advice to the manager.

As the three examples illustrate, the social signals AI agents need to interpret to become socially intelligent can be internal or external to the human-AI relationship itself. That is, AI agents are not only required to interpret the social signals directed directly at them but also those traded between other humans.

While AI models have shown a remarkable capacity to decode and mimic important aspects of human psychology – for example, showing properties resembling theory of mind (Kosinski, 2024) and inductive reasoning (Bhatia, 2023), or the ability to infer and project psychological characteristics (Peters et al., 2024; Peters & Matz, 2024; Serapio-García et al., 2023) – we know much less about their ability to accurately decode social cues in dynamic interactions.

In this paper, we explore whether AI models can accurately decode one of the arguably most important but complex social signals: romantic attraction. Specifically, we test whether Large Language Models (LLMs) can detect external signals of romantic attraction during brief getting-to-know-you interactions between human subjects (i.e., speed-dates).

The speed-dating context offers an interesting testing ground for LLMs for two main reasons: First, speed dates are highly structured and time-limited in format, allowing for a direct comparison across multiple dyads (Finkel et al., 2007). Second, the brevity of interactions makes our test a conservative estimate of LLMs' capacity to interpret the content of social interactions. Likewise, the focus on human-to-human interactions rather than human-AI interactions offers benefits that make our investigation a conservative test of LLMs' ability to detect social cues. First, studying the outcomes of real human-to-human speed dates allows us to observe actual matching behavior in the form of speed dating participants exchanging contact information. While human-AI interactions would have allowed us to test LLMs' accuracy at predicting romantic attraction in their partner, human-to-human exchanges enable us to test their ability to understand more complex dynamics between people using real behavioral outcomes (an arguably more difficult task and hence more rigorous test). Second, focusing on LLMs' ability to decode external signals allows us to more clearly separate their ability to (i) decode social cues of romantic attraction – the focus of this paper – from their capacity to (ii) adequately respond to these social cues. That is, the mere fact that AI agents can influence the nature and content of their conversations with human counterparts could influence their ability to detect social cues. Although our findings cannot directly speak to the ability of LLMs to detect social cues in the context of their own interactions with human counterparts, it is likely that the findings discussed in this paper will generalize to the detection of romantic attraction in direct AI-human interactions.

The analyses presented in this paper aim to establish the general ability of LLMs to predict romantic attraction, to uncover the social cues driving their judgments, and to compare the LLMs' judgements to those made by humans. The last part is critical when considering social judgements as a means for creating a shared understanding of the external social world (i.e., create a sense of shared reality; Higgins et al., 2021). That is, for relational AI agents and their human counterparts to make sense of the social world together they need to not only be accurate in their prediction but also overlap in their judgements as well as the cues they use to derive them.

Specifically, we analyze the transcripts of 964 speed dates to (i) test whether OpenAI's ChatGPT (gpt-4-0613; OpenAI et al., 2024) can accurately predict the success of speed-dates based on short transcripts, (ii) explore how such predictions compare to those made by human observers, and (iii) investigate the behavioral signatures of romantic attraction ChatGPT uses to make its predictions.

**Theoretical Background**

*The Potential of Large Language Models for the Development of Relational Agents*

Interactions between humans and artificial agents – e.g., chatbots or virtual assistants – have become increasingly common. To facilitate a smooth and pleasant user experience that resembles successful human-to-human interaction, a growing proportion of these artificial agents are designed to both engage in natural, free-flowing conversations and to emulate socio-emotional relationships with users (Mariani et al., 2023). By understanding users' needs on a deeper level than mere stated preferences and demands, and by displaying human-like characteristics, such as humor or empathy, for example, social agents aim to move beyond merely transactional exchanges to build trust, rapport, and emotional connections with users (Bickmore & Picard, 2005; Castelfranchi, 1998).

A particularly powerful type of social agents are relational agents, which are designed "to build long-term, social-emotional relationships with their users" (Bickmore & Picard, 2005; R. H. Campbell et al., 2009). In contrast to other types of social agents that primarily adjust to the momentary needs of users, relational agents have a memory of previous interactions – including specific parts of prior conversations – which allows them to reference "shared experiences" and build meaningful relationships over time (R. H. Campbell et al., 2009).

The foundation of any relational agent is their ability to accurately decode social cues as implicit indicators of individuals' thoughts and feelings. While two colleagues gossiping about another coworker might not explicitly express their contempt towards them, for example, their verbal remarks (and non-verbal behavior) likely contain cues to how they feel about that person. Likewise, two people getting to know each other romantically might not explicitly express their attraction with the words "I love you" but their interactions might nonetheless leave cues to their feelings.

While much of the existing literature on relational agents is focused on their ability to decode and respond to the social cues of their immediate interaction partners (Bickmore & Picard, 2005; R. H. Campbell et al., 2009; Mariani et al., 2023), a growing body of research on close relationships in humans highlights the importance of interaction partners making sense of the external world together. Specifically, research on shared reality argues that creating a shared understanding of the socio-physical world around us satisfies important epistemic needs that play a critical role in establishing and maintaining close relationships (Higgins et al., 2021; Rossignac-Milon et al., 2021). For AI agents to become truly relational in nature, they consequently need to be able to decode not only the social cues directed directly at them, but also those traded between humans.

While the concept of relational agents was introduced almost 20 years ago (Bickmore & Picard, 2005), recent advances in Natural Language Processing (NLP) have significantly enhanced the capabilities and widespread deployment of relational agents. Among these developments, the introduction and rapid commercialization of Generative AI in the form of LLMs – such as Open AI's ChatGPT, Anthropic's Claude or Google's Gemini – is undoubtedly the most important one. Trained on vast corpora of textual data, LLMs learn statistical patterns in language, allowing them to both understand natural language and generate novel text that is often indistinguishable from that created by humans (Jakesch et al., 2023). LLMs can answer open-ended questions, summarize content, or emulate natural conversations.

Yet, what makes LLMs a promising tool for relational agents is that they are capable of far more than merely generating text and engaging in free-flowing conversations. As a growing body of research suggests, LLMs possess human-like abilities that allow them to engage in complex social reasoning and interpret the types of social cues needed to understand human interactions beyond what is expressed on the surface. For example, LLMs exhibit properties resembling theory of mind (Kosinski, 2024) the ability to accurately impute the mental states of other entities, and have demonstrated a remarkable ability to detect human emotion and sentiment (Lei et al., 2024; Wang et al., 2023; W. Zhang et al., 2023; Z. Zhang et al., 2024) as well as psychological dispositions such as personality traits (Peters et al., 2024; Peters & Matz, 2024). Similarly, research suggests that LLMs can discern subtle social signals, such as humor (Ko et al., 2024) or sarcasm (Liu et al., 2022), and both identify and explain social norms and norm violations in text, reflecting a capacity to understand implicit social rules (Neuman & Cohen, 2023).

However, despite the promising potential of LLMs to understand (and ultimately mimic social signals) and their growing popularity as relational agents (e.g. Replika or Character.AI), little is known about their ability to interpret more dynamic and complex social interactions that form the foundation of relational agents. As we have outlined above, establishing meaningful long-term relationships requires LLMs them to decode the social meaning of exchanges where social cues develop over time through ongoing interactions. This includes the accurate "interpretation" of whether a conversation between two people or with the agent itself is going well, and whether the human counterpart enjoys the interaction in a way that makes them want to engage in future exchanges.

In this paper we explore the ability of LLMs to detect social cues in dynamic exchanges on speed dates. The domain of romantic attraction – and the context of speed dating in particular – offers a promising test case for evaluating LLMs' capacity for social judgment because it represents a domain where successful inference requires an understanding of not just individual social cues but also their temporal sequence in the broader context of socio-cultural expectations and norms.

*Decoding Social Cues in Romantic Attraction: A Brunswik Lens Model Approach*

We investigate the ability of LLMs to detect social cues in romantic attraction using Brunswik's Lens model (Brunswik, 1956). The model offers a conceptual framework for understanding and systematically analyzing how people – or artificial agents – make judgments about their physical and social environment (see Fig. 1). Whether this process is directed at assessing the quality of a job candidate, predicting the success of a stock on the stock market, or estimating another person's romantic interest, human as well as artificial agents rely on available information, or "cues", to infer a higher-level construct of interest. They might consult a person's CV to scan for indicators of resilience, study prior stock performance in the context of current market trends or look for an overlap in interests that might indicate romantic compatibility. These observable cues function as a lens that mediates the human or artificial agent's perception of the unobservable criterion.

**Figure 1**

*Brunswik lens model applied to the perception of romantic attraction (criterion) by human and artificial agents (judgment) who can access observable indicators (social cues).*

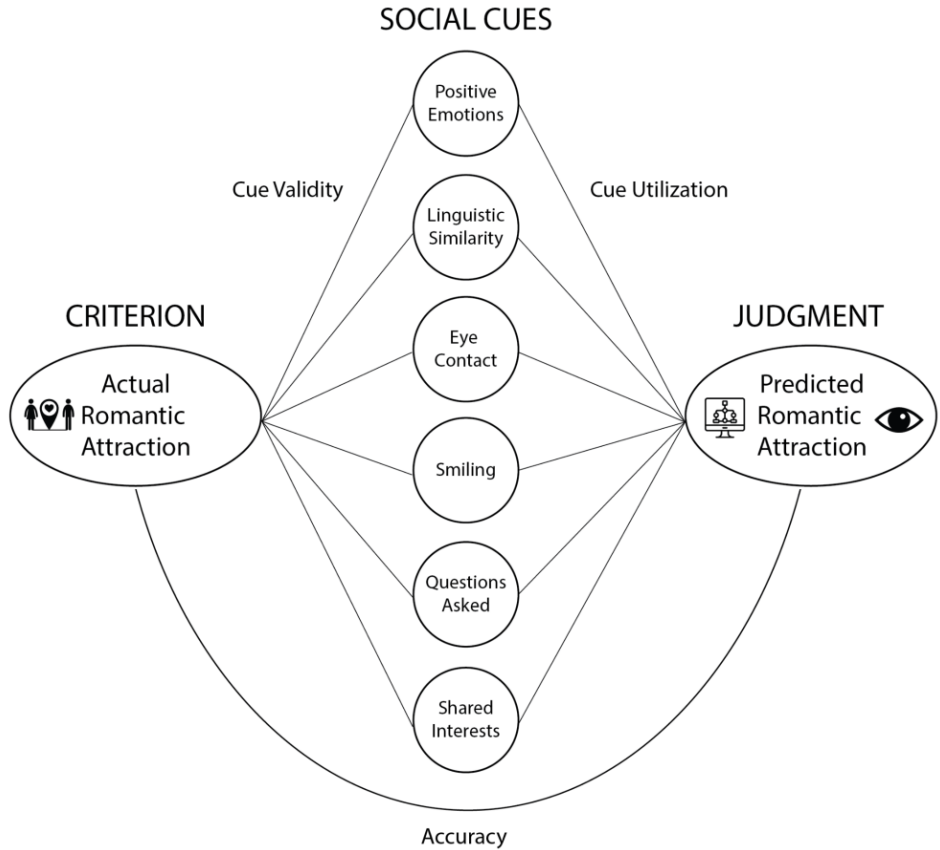

Importantly, the Brunswik lens model introduces several metrics that can be used to probe and evaluate LLMs' ability to accurately decode social cues. First, cue validity represents the relationships between observable (social) cues and the unknown criterion, which is the target of the judgment. In the context of romantic attraction, the observable cue of physical attractiveness, for example, is known to be related to romantic attraction (Back et al., 2011; Eastwick et al., 2023; Montoya, 2008; Walster et al., 1966). Second, cue utilization indicates the extent to which a human or artificial judge uses a particular cue in their judgment of the target. For example, different agents might consider physical attractiveness more or less important in their prediction of romantic attraction, regardless of whether the cue itself is, in fact, valid. Both cue validity – i.e. the presence of valid, observable cues – and cue utilization – i.e., the use of these cues by human or artificial judges – are necessary for accurate social judgments, which are captured by the accuracy metric of the lens model. Specifically, accuracy captures the correlation between the criterion (i.e., romantic attraction) and the prediction made by the human or artificial judge (i.e., predicted romantic attraction).

Our investigation relies on the findings of prior research suggesting that there are at least some valid, observable social cues of romantic attraction (Asendorpf et al., 2011; Back et al., 2011; Eastwick et al., 2023; Joel et al., 2017). Notably, most this existing work has focused on physical attractiveness as a driver of dating choices (Back et al., 2011; Montoya, 2008; Walster et al., 1966) or investigated non-verbal (e.g. laughing, gaze; Grammer et al., 1999, 2000; Renninger et al., 2004) or paraverbal behaviors (e.g. pitch; McFarland et al., 2013; Pisanski et al., 2018) that are unobservable to the purely text-based LLMs deployed in this paper. However, there is at least some evidence for verbal indicators of romantic attraction in the context of online dating (Lee et al., 2019) and existing romantic relationships (Bowen et al., 2017; Karan et al., 2019; Simmons et al., 2005). Using the structural analysis of social behavior coding scheme (SASB; Dai & Robbins, 2021) for example, Eastwick and colleagues (2010) identified several verbal indicators of smooth versus awkward dates. Participants on dates that were rated as smooth tended to speak more warmly and were more other-focused than those on awkward dates (Eastwick et al., 2010). Similarly, romantic attraction has been linked to linguistic style matching (i.e., the extent to which people are similar in how they talk with one another; Ireland et al., 2011), as well as the frequency by which speed dating participants asked follow-up questions (Huang et al., 2017). Taken together, existing research suggests that there are at least some valid verbal indicators of romantic attraction (i.e., some degree of cue validity) LLMs could draw on when judging the social dynamics of speed dates.

However, it remains an open question whether LLMs indeed utilize these – or other currently unknown — cues to romantic attraction in their social judgments. On the one hand, the ability of LLMs to process and analyze vast amounts of language data reflecting the human experience across a wide range of domains (e.g., social media posts, news articles, stories, and cultural explorations) makes them ideal candidates for the detection of social patterns. If LLMs are indeed able to detect cues of romantic attraction, this would not only support their potential for relational agents but also offer a promising path for better understanding romantic attraction itself. For example, it is possible that LLMs identify cues that are valid indicators but have been overlooked by previous investigations, which have more narrowly focused on theoretically meaningful, pre-defined cues.

On the other hand, most current LLMs still lack the ability to process crucial non-verbal and paraverbal cues—such as facial expressions, body language, tone of voice, and timing—that often play an important role in romantic interactions. Additionally, there is a risk that AI models merely perpetuate stereotypical patterns of romantic attraction that form the foundation of their training and overlook more idiosyncratic expressions of romantic preferences and expressions.

In this paper, we aim to empirically test LLMs' ability to detect social cues of romantic attraction Specifically, our analyses aim to answer three interrelated questions. Can LLMs like ChatGPT predict romantic interest from brief social interactions between humans? How does their predictive accuracy compare to that of human observers? Which linguistic cues do LLMs (and human judges) rely on in their predictions, which among these cues are valid, and which cues might LLMs be missing out on?

## Methods

**Participants and Procedure**

The original dataset contained 1,039 speed dates from 187 undergraduate students (93 women and 94 men; age = 19.6±1.2 years old) who attended one of eight speed dating events at a large Midwestern University in 2007. Participants were recruited via flyers and e-mails and engaged in approximately 12 speed dates, each lasting four minutes and consisting of a man and a woman (see **Fig. 2** for an overview of the Study Design and Supplemental Material A in the SI for recruitment materials). Immediately following each speed date, participants evaluated the experience (see the Methods section for more details and refer to Finkel et al., 2007, for the original publication of the dataset). The data collection for this study was approved by the Ethics Committee of [Masked for Review] University (Protocol #1343-019).

**Figure 2**

*Overview of study design.*

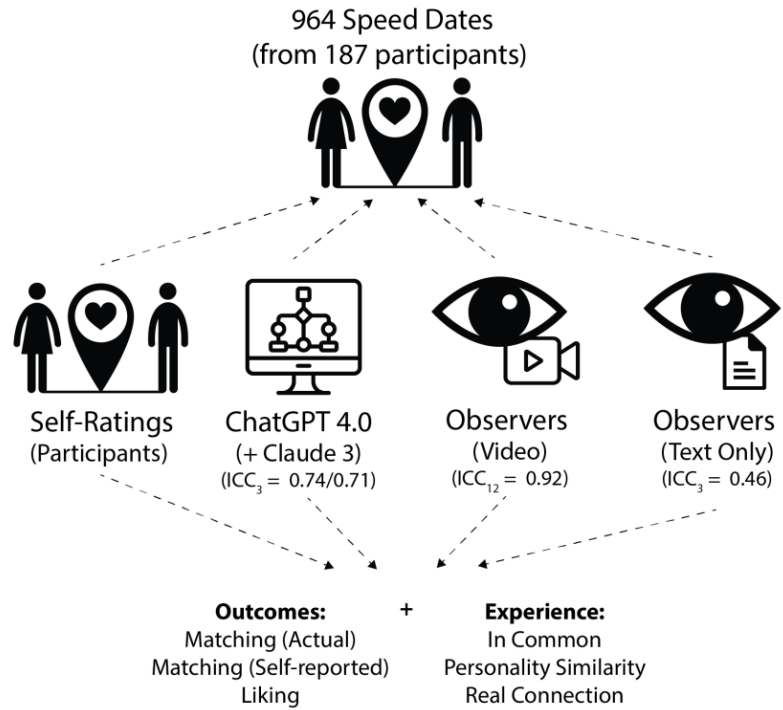

*Note.* Speed dates were independently evaluated by four judges: speed date participants, ChatGPT (and Claude 3 in supplementary analyses supporting generalizability), human observers with access to video recordings of the speed dates, and observers with access to speed date transcripts only. All judges provided information on both the (predicted) speed dating outcomes and experiences. ICC = Intraclass Coefficient (measure of inter-rater reliability).

All speed-dates were recorded on video and transcribed by human professionals. We were able to match 964 of the transcripts based on their unique identifiers in the self-report data and file names of the transcripts, which included an average of 864±147 words per date. The data and code needed to replicate our findings are openly available on the project's OSF page (view-only for peer review purposes).

**Measures**

The speed dates were evaluated by four independent sources: (i) the speed-daters immediately after the interaction; (ii) ChatGPT (gpt-4-0613), which analyzed the written transcripts of the speed-dates (the Supplementary Information also reports the findings for judgements made by Anthropic's Claude3, supporting the generalizability of our findings to other LLMs), (iii) eighteen judges (of which 12 raters scored every item), who watched the speed-dating videos; and (iv) four judges who read a random subset of 500 speed-dating transcripts without access to the videos.

All four sources rated the same set of five statements using a 9-point Likert scale (see Supplemental Materials B, C, and D in the SI for the specific items, the distributions of judgments, and ICCs). Two of the statements captured speed-dating *outcomes*: (i) participants' level of liking for their partner (e.g., ", and (ii) the likelihood of saying "yes" to a potential follow-up date (e.g., "I am likely to say yes to my interaction partner"). The other three captured speed-daters' *experiences* as potential mechanisms: participants' beliefs that they (i) had "a lot in common" with their partner, (ii) had "similar personalities," and (iii) experienced "a real connection." ChatGPT and the human judges reading the transcripts rated the outcomes at the dyad level (e.g. "On a scale from 1-9, how likely do you think the interaction partners will say "yes" to each other and exchange contact information?"). In contrast, both participants and the human judges with access to the video recordings rated them at the level of individual participants; To score the speed-date as a whole, we averaged those individual ratings across both interaction partners.

While interrater reliabilities for the human judges with access to the speed dating videos were high (average ICC = 0.92 across the two speed dating outcomes and experiences), they were significantly lower for the human judges who only read the transcripts (average ICC = 0.46).

Notably, the interrater reliabilities associated with the ratings produced by ChatGPT (and Claude 3 in the SI) can be influenced by adjusting the model's temperature, a parameter that determines how similar the model's responses are across different rounds (OpenAI et al., 2024). For the purpose of this study, we kept the models' standard settings (temperature=1) to test their baseline performance. The average agreement across the three independent API queries for ChatGPT was ICC=0.74 (average ICC = 0.71 for Claude 3). On average, the ratings of ChatGPT and Claude 3 were correlated by 0.71.

After the speed-dating event concluded, participants indicated whether they wanted to exchange contact information with each partner for a potential follow-up date. This mutual signal of interest or "matching" served as a third, objective speed-dating outcome, and our main metric of romantic attraction. Among all the speed dating dyads, 23% matched by exchanging contact information.

## Results

**Can ChatGPT predict speed dating outcomes and experiences?**

Our primary analysis tested whether LLMs can predict the objective matching outcome – participants exchanging contact information – by comparing ChatGPT's predictions to the criterion of actual matching (i.e., accuracy in the Brunswik lens model). To develop a better understanding of the unique capabilities and insights afforded by LLMs, we evaluate their accuracy vis-à-vis those obtained by human judges, which include both the self-reported intentions of the speed-dating participants themselves as well as the post-hoc ratings made by human judges based on the speed dating transcripts or videos (**Fig. 3A)**. The zero-order correlations of all variables and the equivalent results for Anthropic's Claude 3 (claude-3-opus-20240229) can be found in Supplemental Materials E and F in the SI.

Although ChatGPT's predictive accuracy of actual matching was relatively low ($r=0.12$, $p<.001$), the LLM performed on par with the human raters who had access to the same information (i.e., transcripts only; $r=0.13$, $p=.003$). As expected, the human coders who could also draw on the non-verbal cues observable in the video recordings outperformed both ChatGPT and their human counterparts with less information ($r=0.31$, $p<.001$).

Notably, ChatGPT successfully predicted matching above and beyond participants' self-reported intentions ($B=0.19$, $SE=0.09$, $z=2.17$, $p=.030$). In contrast, we did not observe incremental predictive power for the ratings of the human judges who had access to transcripts only ($B=0.03$, $SE=0.16$, $z=0.19$, $p=0.853$). This suggests that ChatGPT observed unique social cues in the speed dating transcripts that were neither accessible to the speed-dating participants themselves nor the coders who evaluated the same transcripts (note that human coders in the video condition did provide incremental predictive power: $B=0.46$, $SE=0.10$, $z=4.62$, $p<.001$.).

Next, we examined the ability of ChatGPT and human judges to predict participants' judgments of romantic attraction at the end of the speed-date. ChatGPT predicted participants' ratings with correlations ranging between $r = 0.13$ to $r = 0.23$ (all $p<.001$; **Fig. 3B**), even after accounting for word count. The correlations were markedly lower than those made by human judges in both conditions (average correlation for observers with access to videos $r = 0.46$, and transcript only $r = 0.33$), suggesting that human judges reading the transcripts are better at predicting participants' self-reported experience of the speed-date but not their actual commitment to exchanging contact information (actual matching). Notably, the relative levels of accuracy across predicted metrics appeared largely consistent across all judges, with participants' more proximal experiences being easier to predict (in particular, whether they had a lot in common) than the more distal, overall evaluations of the speed dates.

**Figure 3**

*Correlations between the speed-dating ratings of different evaluators.*

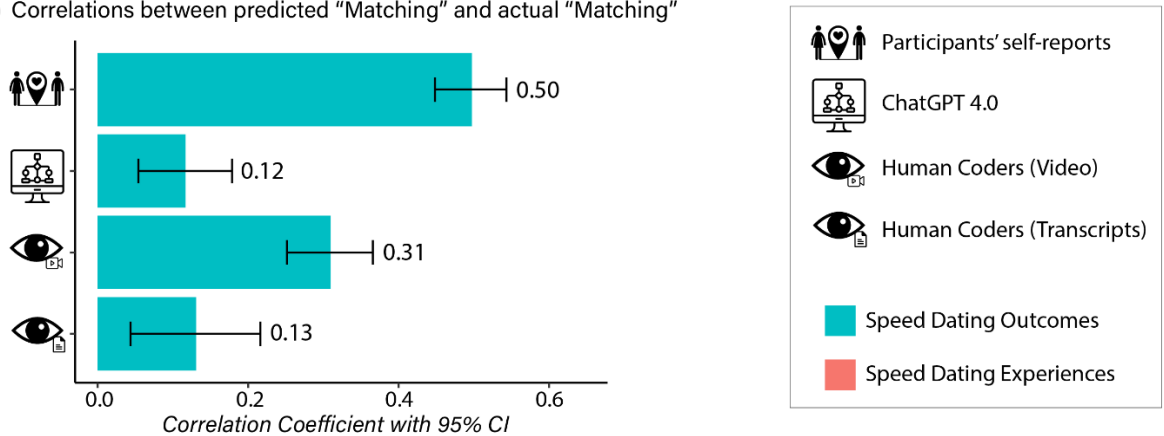

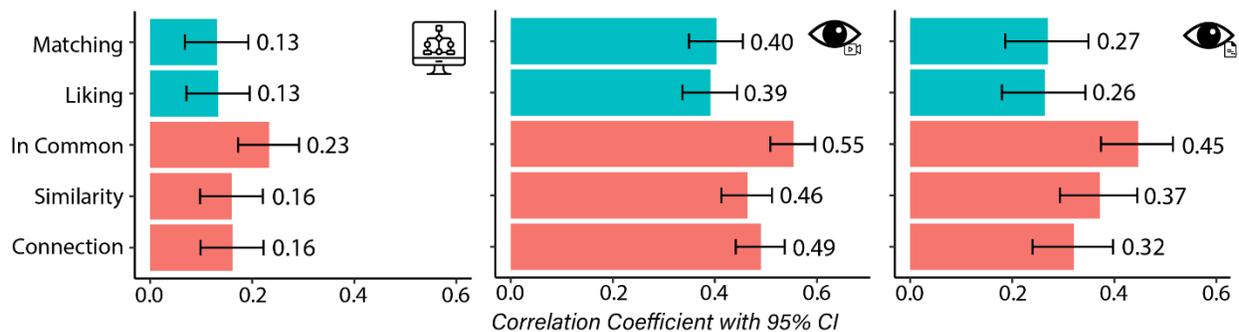

Note. A) Point-biserial correlations between the behavioral speed-dating outcome of matching and the predictions made by different judges of whether participants would exchange contact information. B) Pearson correlations between participants' self-reported outcomes and experiences at the end of the speed-date and the predictions made by ChatGPT and the human judges.

**Do ChatGPT's predictions overlap with those made by human judges?**

In addition to comparing the predictive accuracies of ChatGPT and human raters, we also investigated the extent to which ChatGPT and human judges relied on similar cues when making their predictions. As the correlations in **Fig. 4** show, there was meaningful overlap between the predictions of ChatGPT and the human judges (mean $r$ = 0.29 across all metrics and conditions). The fact that these correlations were higher than those observed with participants' self-reported scores (**Fig. 3B**) suggests that both ChatGPT and human judges utilize cues that are commonly considered markers of successful interactions but are not empirically associated with actual speed-dating outcomes or experiences (see the following results section for more details).

**Figure 4.**

*Correlations between ChatGPT's predictions and the ratings made by human judges.*

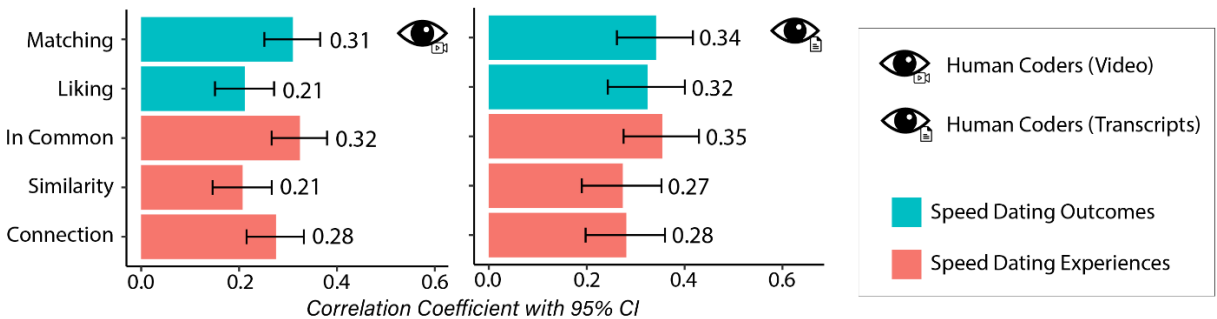

Although the predictions of ChatGPT and human judges overlap, they were not redundant in predicting speed dating outcomes and experiences. For at least some of the metrics, both ChatGPT and human judges offered significant predictive power when added simultaneously into a linear regression model (notably, the effects of ChatGPT became non-significant when added with both, text-based and video-based, human judgments at the same time).

**Which textual features and conversational dynamics are driving the predictions of ChatGPT, and are these cues valid indicators of romantic attraction?**

To better understand how ChatGPT arrived at its predictions and to test whether the predictions were merely driven by the linguistic features of the conversation (e.g. emotional tone) rather than conversational dynamics, we conducted two sets of additional analysis: First, we mapped ChatGPT's predictions against established markers of linguistic styles. Specifically, we used LIWC (Boyd et al., 2022) – a widely used, dictionary-based text analysis tool – to categorize and quantify the psychological, emotional, and cognitive aspects of the speed dating transcripts along 117 validated dictionaries. Second, we prompted ChatGPT to explain each of its predictions and subsequently asked it to synthesize these explanations in a parsimonious taxonomy of romantic attraction markers. In both cases, we applied a Brunswik lens approach to compare the cues utilized by the model (i.e., the language dimensions correlated with ChatGPT's predictions of attraction) to the cues that were actually valid indicators of romantic attraction (i.e., exchange of contact information by participants).

*LIWC Mapping*

To explore the linguistic features associated with ChatGPT's predictions as well as actual matching, we extracted all 117 available LIWC dimensions for each of the transcripts. The dimensions cover a wide range of categories, such as linguistic dimensions (e.g., pronouns, verb tense), psychological constructs (e.g., affective processes, social concerns, cognitive processes), and personal concerns (e.g., work, money, health). We subsequently correlated the LIWC scores for each transcript with the predicted matching scores as well as actual matching outcome (see OSF for all correlations).

**Table 1** displays the fifteen LIWC dimensions that showed the highest absolute correlations with GPT's predicted likelihood of saying "yes.", capturing ChatGPT's cue utilization independent of accuracy. The examples suggest that GPT indeed relied on the affective valence of conversations but also highlight the importance of other dimensions unrelated to emotional valence (e.g., high levels of certitude and a reduced focus on the present). In addition, the table captures cue utilization of human judges for the same dimensions as well as cue validity (i.e., correlation with the actual outcome). As the comparison between these different variables suggests there is a strong overlap between the cues utilized by ChatGPT and Humans, but not always high levels of accuracy (i.e., utilized cues are not always valid indicators). In some instances, the utilized cues are not only non-predictive of actual matching but, in fact, point in the opposite direction. For example, both ChatGPT and humans consider negations and the use of common verbs a negative indicator of romantic attraction, while both are indeed positively associated with actual matching.

**Table 1.**

*LIWC dimensions that were most strongly correlated with ChatGPT's predictions of romantic attraction (i.e., cue utilization) alongside the cue utilization of human judges and cue validity.*

| Dictionary | Examples | Cue Utilization (ChatGPT) | Cue Utilization (Human) | Cue Validity |
|---|---|---|---|---|
| Negations | Not, no, never, nothing | -0.29 | -0.06 | 0.09 |
| Positive Tone | Good, well, new, love | 0.28 | 0.19 | 0.05 |
| Emotional Tone | Degree of positive (negative) tone | 0.26 | 0.16 | 0.03 |
| Affect | Good, well, happy, hope | 0.26 | 0.20 | 0.06 |
| Positive Emotion | Good, love, happy, hope | 0.20 | 0.31 | 0.09 |
| Word Count | Total word count | 0.20 | 0.36 | 0.13 |
| Emotion | Degree of positive (negative) emotion | 0.19 | 0.28 | 0.08 |
| Auxiliary Verbs | Is, was, be, have | -0.18 | -0.12 | 0.06 |
| Differentiation | But, not, if, or | -0.17 | -0.11 | -0.01 |
| Analytical Thinking | Metric of logical, formal thinking | 0.16 | 0.024 | -0.07 |
| Present Focus | Is, are, I'm, can | -0.15 | -0.14 | 0.02 |
| Common Verbs | Is, was, be, have | -0.15 | -0.06 | 0.09 |
| 3rd Person Plural | They, their, them, themsel* | -0.15 | -0.01 | 0.06 |
| Social Referents | You, we, he, she | -0.15 | -0.03 | 0.05 |
| Certitude | Really, actually, of course, real | 0.14 | 0.08 | -0.003 |

Note. Cue Utilization = correlation between judgement and cue, Cue Validity = Correlation between actual matching and cue

To further support the proposition that ChatGPT did not merely pick up on readily available content features, we trained a supervised model predicting ChatGPT's judgments of mutual attraction (i.e., the exchange of contact information) from a linear combination of the 117 LIWC dimensions using the following procedure: We randomly selected 50% of the data (training dataset with 499 speed dates) to train a LASSO model, tuning the model parameter lambda in a 10-fold cross-validation. We subsequently applied the model to the remaining 50% of the data (testing dataset with 461 speed dates). Then out-of-sample validation of the model on the testing data suggests that LIWC accounted for only ~12% of the variance in ChatGPT's predictions ($r=0.35$). This highlights that a substantial amount of the variance in the predictions made by ChatGPT cannot be captured by a linear combination of the relatively comprehensive set of content dimensions included in the LIWC dictionaries.

To explore whether the cues utilized by ChatGPT overlap with those correlated with actual matching, we divided the 177 LIWC dimensions into four buckets: True positive, false positive, false negatives and true negatives. The results are displayed in **Table 2**. The first column (True Positives) includes all dimensions that were significantly associated with both predicted and actual matching. In other words, the dimensions shown in this column represent the linguistic features that were accurately identified by ChatGPT as valid indicators of romantic attraction, with correlations being both significant and pointed in the same direction. The second column (False Positives) includes all dimensions that were significantly associated with predicted but not actual matching. That is, these dimensions reflect cues that were considered important by ChatGPT but were not actually indicative of participants exchanging contact information. Finally, the last column (False Negatives) includes dimensions that were significantly associated with actual but not predicted matching. As such they indicate linguistic features that were important drivers of romantic attraction but overlooked by ChatGPT. All dimensions are sorted by effect size with the highest absolute correlations displayed at the top of the column. The remaining dimension – true negatives which represent all dimensions that were neither predicted to be important nor actually predictive of matching – can be found in Supplemental Material G of the SI.

As the breakdown of Table 2 suggests, there was only a relatively small set of valid linguistic cues utilized by ChatGPT (i.e., true positives). These included both the emotional tonality of the conversation, overall word count as well as references to social behaviors. In contrast, there were a larger number of invalid cues ChatGPT utilized for its predictions (i.e., false positives) as well as valid cues it failed to consider (i.e., false negatives).

**Table 2**.

*Categorization of LIWC dimensions into cues that were utilized and valid (true positives), utilized but not valid (false positives) and valid but not utilized (false negatives).*

| True Positives | False Positives | False Negatives |
|---|---|---|
| Positive Emotion (+)(H) | Positive Tone (+)(H) | Work (-) |
| Word Count (+)(H) | Tone (+)(H) | Lifestyle (-) |
| Emotion (+)(H) | Affect (+)(H) | Money (-)(H) |
| Tech (-) | Differentiation (-)(H) | Exclamation points (+) |
| Social Behaviors (+)(H) | Present Focus (-)(H) | Motion (+)(H) |
|  | Social Referents (-) | 3rd Person Singular (+) |
|  | Certitude (+) | Question Mark (-) |
|  | Future Focus (-) | Past Focus (+)(H) |
|  | Cognition (-) | Personal Pronouns (+) |
|  | Insight (-) | Allure (+)(H) |
|  | Feeling (+) | Substances (+) |

|  |  |
|---|---|
| Causation (-) | Affiliation (+)(H) |
| Apostrophes (-) | |
| Function Words (-) | |
| Conversational (+) | |
| Cognitive Processes (-) | |
| Sadness (+) | |
| Female References (-) | |
| Home (-) | |
| Want (+) | |

Note. (+) and (-) = Positive or negative correlation with ChatGPT's prediction (cue utilization), (H) Cue is also utilized in the same direction (or not utilized in the case of false negatives) by human judges.

Notably, many of the "mistakes" made by ChatGPT were also observed in human raters (all indicators marked with an H in Table 2). For example, we found that both human raters and ChatGPT consider a positive tone as a strong indicator for a successful match (i.e., high correlations between the predicted matching ratings and the respective LIWC dimensions), even though this cue is not strongly associated with actual matching. That is, the correlations between positive tone and the predicted matching scores of ChatGPT and human raters were $r = 0.28$ and $r = 0.19$ respectively, with positive emotions ranking among the top five predictors in both cases. However, positive emotion was only correlated at $r = 0.05$ with actual matching, ranking number 44 among all 117 predictors. In contrast, references to money were one of the strongest (negative) predictors of actual matching, but neither ChatGPT nor the human raters considered them important.

Taken together, the above analyses suggest that ChatGPT's judgments of romantic attraction do not simply rely on readily observable linguistic features, but instead involve the interpretation of conversational dynamics. To further explore the specific conversational dynamics utilized by ChatGPT and relate them more directly to its judgements, we conducted a series of additional analyses that rely on the model's capacity to explain its predictions.

*ChatGPT-based Explanations*

To further probe the conversational cues underlying ChatGPT's predictions, we prompted the model to explain the predictions it made through a zero-shot chain-of-thought process (Kojima et al., 2022). Specifically, we prompted ChatGPT: "How likely do you think the interaction partners will say "yes" to each other and exchange contact information? Let's think step by step." (see our OSF page for all generated explanations). A manual inspection of the explanations revealed that ChatGPT referred to many of the same indicators of romantic attraction previously identified in the literature (e.g. smoothness of conversation; Eastwick et al., 2010, 2024). Instead of imposing a pre-existing taxonomy on the explanations generated by ChatGPT, we asked the model to "*Create a parsimonious taxonomy of explanation categories that captures recurring explanations and are psychologically meaningful*" (see Supplemental Material H in the SI for full details). Given the limited number of tokens per API query, we limited ChatGPT to the first 100 explanations.

The proposed taxonomy included eight broad categories (e.g., Conversation Flow and Engagement), and 19 secondary indicators (e.g., Smoothness of Conversation, Mutual Engagement, Humor and Playfulness). Before proceeding with additional analyses, we excluded three categories we reasoned would be difficult to rate reliably based on textual data alone (i.e., External Factors and Contextual Influences, Internal States and Perceptions, Non-Verbal and Unspoken Cues). Notably, the inclusion of these categories in the proposed taxonomy could be explained in two ways: First, it is possible that ChatGPT complemented its analysis of the specific with its generic representation of romantic attraction. Second, many explanations mentioned the categories and its indicators as important drivers that could *not* be considered given the purely textual input (e.g. body language). Although these explanations explicitly highlight the absence of these cues from the transcripts, it is possible that ChatGPT nonetheless included them in the creation of its overall taxonomy. The final taxonomy with primary categories and secondary indicators is displayed in Table 3 (see Supplemental Material I in the SI for additional explanations of the indicators and excluded categories).

**Table 3**

*Bottom-up taxonomy of indicators of romantic attraction generated by ChatGPT*

| Primary Category | Secondary Indicator |
|---|---|
| Conversation Flow and Engagement | Smoothness of Conversation |
|  | Mutual Engagement |
|  | Humor and Playfulness |
| Shared Interests and Values | Common Interests |
|  | Aligned Values and Goals |
| Emotional and Personal Connection | Emotional Depth |
|  | Reciprocity of Disclosure |
|  | Flirtation and Romantic Signals |
| Perceived Compatibility and Future Potential | Perceived Compatibility |
|  | Expression of Future Intentions |
|  | Social Compatibility |
| Conversational Challenges | Moments of Tension or Disagreement |
|  | Mismatch in Interests or Values |

As a first step in exploring the content of ChatGPT's explanations, we calculated the relative frequencies by which each of the indicators was mentioned as part of the explanations (see Supplemental Material J in the SI for details on how we extracted the frequencies). As Figure 5 shows, mutual engagement, common interests, perceived compatibility and smoothness of conversation were among the most commonly referenced explanations, followed by reciprocity and humor.

**Figure 5**

*Relative frequencies of indicators of romantic attraction mentioned in ChatGPT's explanations of its predictions across all 964 speed dates.*

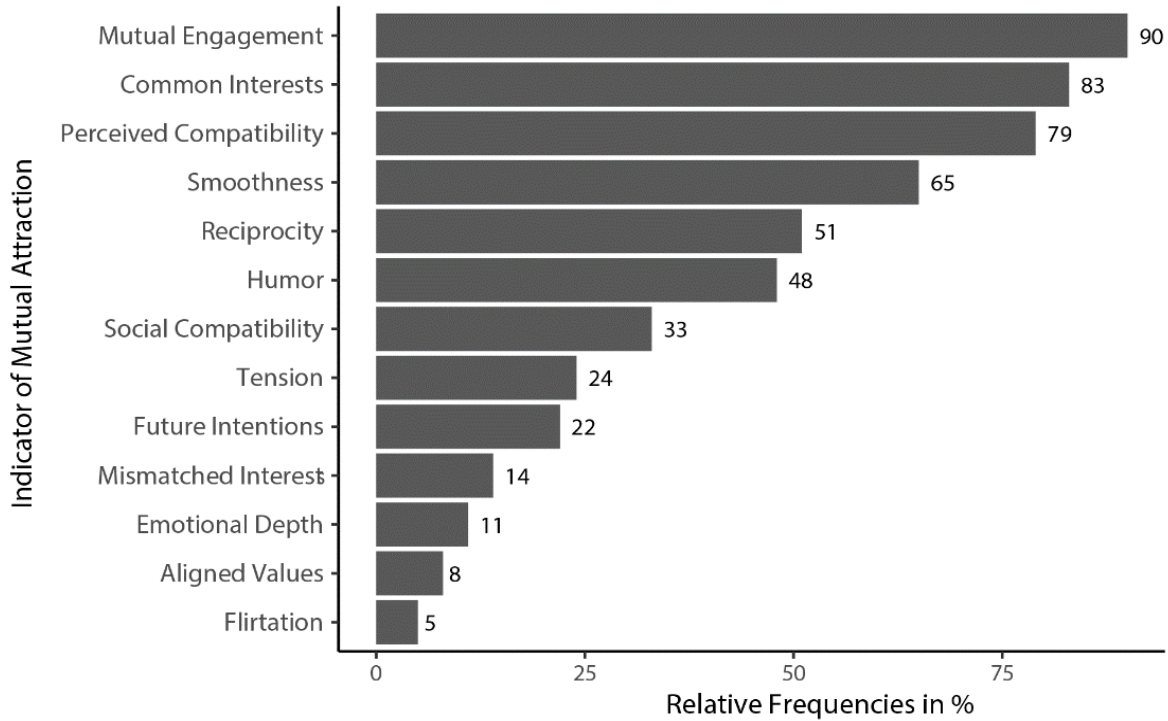

Next, we tested whether the explanations provided by ChatGPT were indeed related to its predictions of romantic attraction as well as actual matching. For this purpose, we prompted ChatGPT to rate each transcript on the thirteen indicators as well as the likelihood of matching (e.g. "*On a scale from 1-9, how smoothly did the conversation progress without awkward pauses or forced topics?*", see Supplemental Material K in the SI for the full prompt). As before, we repeated this exercise three times and calculated the average rating across all three rounds (average ICC = 0.61). Figure 6 displays the correlations between the thirteen indicators of romantic attracted and predicted matching scores (dark grey bars). Supporting the validity of ChatGPT's explanations, all except one indicator (i.e., Mismatched Interests) were significantly correlated with its predictions of matching.

Similar to the Brunswik lens approach applied to the LIWC dimensions, we further tested which of the utilized cues were indeed valid predictors of actual matching. To do so, we calculated the correlations between indicator ratings and actual matching (i.e., participants exchanging contact information). As Figure 6 shows, most indicators used by ChatGPT (i.e., 10 out of 13) were significantly associated with actual matching (light grey bars). These include Humor, Flirtation, Future Intentions, Perceived Compatibility, Common Interests, Social Compatibility, Emotional Depth, Aligned Values, Reciprocity and Mutual Engagement, in order of importance. Notably, when submitted to a linear logistic regression predicting actual matching, all ChatGPT-rated indicators combined explain approximately twice as much variance (based on the adjusted R-squared) as the direct matching prediction made by ChatGPT. This highlights the potential value of integrating theoretical depth into machine-based predictions.

**Figure 6**

*Correlations between indicators of romantic attraction as rated by ChatGPT as well as predicted matching (dark grey bars) and actual matching (light grey bars).*

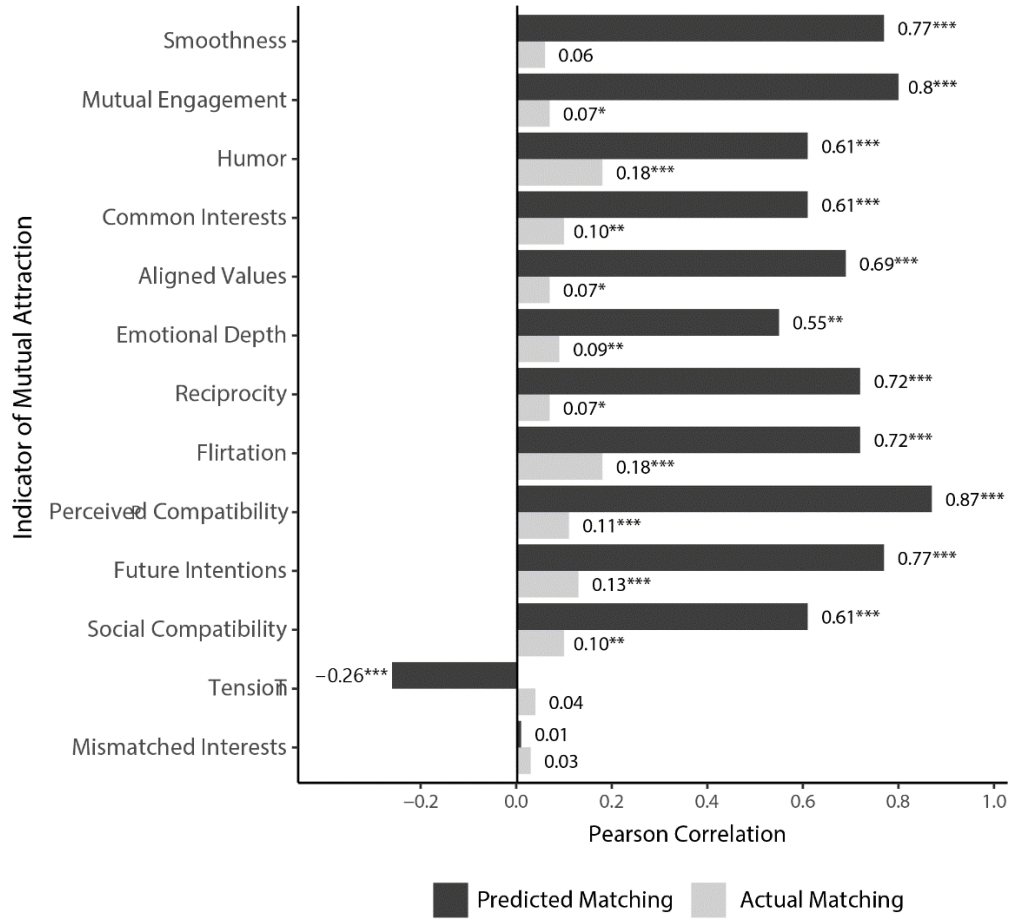

Finally, we tested the indicators as generated and rated by ChatGPT against the benchmark of human judgements. That is, we correlated the indicator ratings with the matching judgements made by the human observers with access to the speed dating videos as well as transcripts only. Mirroring the previous findings in overlap between ChatGPT-based predictions and human ratings (see Figure 4), the correlations displayed in Figure 7 suggest that the human judges implicitly used many of the same cues used by ChatGPT.

**Figure 7**

*Correlations between indicators of romantic attraction as rated by ChatGPT as well as human-coded matching of observers with access to the full video (dark grey) and transcripts only (light grey).*

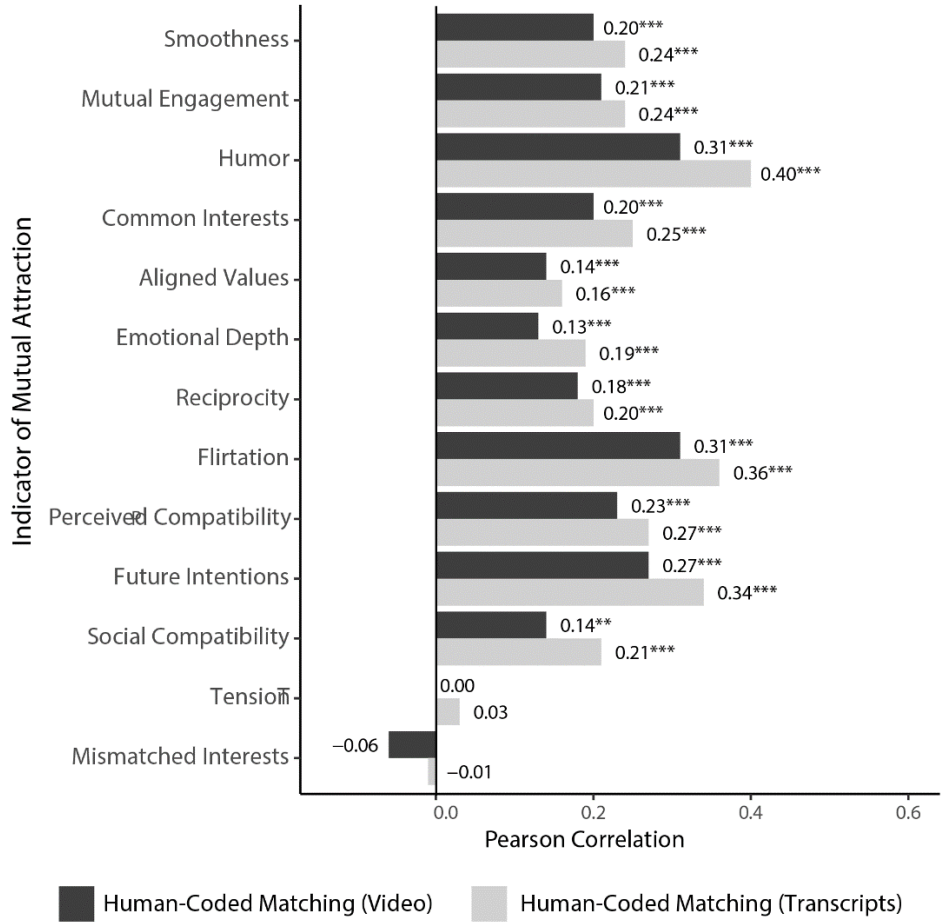

**Discussion**

Taken together, our findings suggest that LLMs can detect some of the conversational nuances that lead strangers to "click" during romantic getting-to-know-you interactions. The accuracies were modest in magnitude ($r$=0.12-0.23). Yet, when predicting whether two people will eventually exchange contact information, the predictions made by ChatGPT were not only on par with those of human judges who had access to the same information (transcripts) but incremental to participants' own ratings at the end of the speed date. In addition, they were moderately correlated with the predictions made by human raters ($r$=0.21-0.35), suggesting that LLMs might pick up on the same cues, even if those cues are not necessarily valid (see Brunswik's lens model; Brunswik, 1956).

Importantly, our findings help explain why the predictive accuracies remain relatively low. As our comparison between utilized and valid cues (see Tables 1 and 2 as well as Figure 6) suggests, there are several mismatches where ChatGPT – similar to human observers – either uses cues that are invalid or overlooks cues that are valid. A better understanding of these mistakes could eventually help improve predictive accuracy in a modeling approach that leverages both theory development as well as data-driven pattern recognition.

In addition, our findings highlight the challenging nature of predicting romantic attraction. As our results suggest, ChatGPT – as well as human raters – were generally more accurate at predicting subjective attraction ratings than actual matching outcomes. This finding aligns with previous research showing that momentary feelings of attraction do not always translate into concrete dating decisions (Asendorpf et al., 2011), underlining the complexity of romantic attraction judgments (as well as their underlying decision-making processes on the actor side) as a multistep and multifaceted process. Instead of being a monolithic concept, romantic attraction involves both immediate subjective experiences and additional context-driven considerations which highlights the importance of differentiating between the prediction of "in-the-moment" romantic attraction and the more challenging task of predicting the actual pursuit of a romantic partnership.

**Contributions and Future Research**

Our findings offer three core contributions that both support LLM's potential to act as relational agents and add to our understanding of romantic attraction as part of the close relationships literature. First, we provide novel evidence for the remarkable abilities of LLMs to solve tasks that were previously considered to be the exclusive domain of human agents (Bai et al., 2023; Kosinski, 2024; Orrù et al., 2023; Peters & Matz, 2024). Much of the existing literature on AI models and artificial agents has focused on interpreting and mimicking the behavior of individuals, such as predicting people's personality from their social media profiles or short free-flow conversations (Peters et al., 2024; Peters & Matz, 2024), creating persuasive content that is customized to the psychological characteristics of their counterparts and often more persuasive than that generated by humans (Bai et al., 2023; Matz et al., 2024), or solving tasks that require theory of mind (Kosinski, 2024) as well as inductive reasoning (Bhatia, 2023). Our work expands this literature by highlighting the ability of LLMs to capture social dynamics playing out between individuals, and as such supports a basic premise underlying relational artificial agents: The ability to make sense of social cues exchanged between artificial agents and human counterparts as well as those traded between humans.

The latter – i.e., interpreting social cues exchanged between humans that are external to the actual agent-human relationship – is necessary for relational agents to create a sense of shared reality with their human partner. As a growing body of research shows, making sense of the external world – not just one's own relationship – together satisfies both key epistemic needs, such as certainty and meaning in life, and key relational needs, such as trust and relationship satisfaction (Elnakouri et al., 2023; Enestrom et al., 2024; Rossignac-Milon et al., 2021; Rossignac-Milon & Higgins, 2018). Supporting the potential of LLMs to act as relational partners, our findings suggest that they are not only able to interpret complex social cues with some degree of accuracy, but that their judgments overlap substantially with those made by humans (see Fig. 4 and Fig 6). This suggests that LLMs might be able to create a sense of shared reality with their human partners even when both are inaccurate in their social judgments.

Importantly, the ability to interpret social cues and align their judgement with those of humans is only the first of many necessary steps towards developing successful long-term relationships that feel "real" and indistinguishable from those developed between humans. That is, in addition to interpreting social cues, relational agents also need to adequately respond to these cues to build rapport and establish trust. While prior work hints at the ability of LLMs to do so – for example, spontaneously adjusting to the preferences of their conversation partners by being compliant in response to polite and emotionally framed prompts (Vinay et al., 2024) – this step needs to be studied more extensively in the context of meaningful long-term relationships

Second, our findings offer novel insights into the verbal indicators of romantic attraction, supporting the proposition that verbal exchanges alone – without the observability of nonverbal or paraverbal cues – can reveal meaningful signals of romantic interest (i.e., both LLMs and human raters were able to predict actual matching at above-chance levels based on speed dating transcripts alone). While most prior research has focused on static cues such as physical attractiveness (Back et al., 2011; Montoya, 2008; Walster et al., 1966), or nonverbal cues such as laughing or the gazes of dating partners (Grammer et al., 1999, 2000; Renninger et al., 2004), our research highlights the importance of more dynamic conversational features such as mutual engagement, shared interests or the reference to social activities. In doing so, our findings also go beyond existing work on verbal indicators of romantic attraction which has either focused on very specific mechanisms of language-based interaction (e.g. language style matching; Ireland et al., 2011) or the overall quality of the interactions (Sprecher & Duck, 1994).

Finally, our research contributes to a better understanding of how LLMs (and humans) integrate available social cues into judgments about romantic attraction. Following a Brunswik lens approach, we showed how the predictions made by ChatGPT and human judges rely on common cues that are not necessarily valid predictors of romantic attraction. That is, while some of the cues utilized were indeed related to participants exchanging contact information after the speed date (e.g. references to emotions and social behaviors as well as perceived humor, flirtation or social compatibility; see the first column of Table 2 as well as Fig. 6), many of the cues utilized were not empirically related to matching (e.g., markers of certitude, future focus and insight as well as perceived smoothness or tension; see the second column of Table 2 as well as Fig. 6). In addition, both ChatGPT and human coders overlooked valid cues that were indicative of matching (e.g. references to money and work (negative); see the third column of Table 2). Our findings align with existing research, demonstrating how people's lay theories of what inspires romantic attraction often differ from what actually inspires attraction (Eastwick et al., 2024).

**Limitations**

It has not escaped our notice that our study has several important limitations that should be addressed by future research. First, as speed-dates are (by definition) brief, it is possible that LLMs might yield more accurate judgments when observing longer initial interactions. This is particularly true as speed dates might be rather similar in structure and content and remain more superficial than other types of conversations due to their introductory nature (i.e., most people will start with brief introductions and small talk). Future research should investigate different types of social interactions (e.g. couples' text messages) and explore whether the accuracy of predictions can be improved by analyzing longer text excerpts.

Second, our analyses focused exclusively on short-term success metrics, such as self-reported liking or the exchange of contact information. Future research should investigate the ability of LLMs and human judges to predict medium to longer-term success metrics (e.g., the likelihood of entering into a relationship and relationship satisfaction) and explore how attraction judgements change during the early stages of a developing relationship. A particularly interesting period to study might be the transition from initial interest to sustained romantic commitment (L. Campbell & Stanton, 2014), which requires romantic partners to incorporate new information and reevaluate prior impressions in response to changing relational contexts (Eastwick et al., 2011). Akin to the romantic partners themselves, external judges (i.e. artificial agents or humans) might need to update their use of different social cues across different stages. For instance, while an initial attraction judgment might be influenced by observable surface traits, later decisions – such as whether to pursue exclusivity or deepen emotional intimacy – may depend on different information or a more nuanced integration of behavioral patterns and compatibility cues (Asendorpf et al., 2011; L. Campbell et al., 2006; Eastwick et al., 2011). From a lens-model perspective, this means that individuals not only rely on new information but also reinterpret and weigh existing cues as the relationship progresses. What seemed attractive in a first encounter may take on a different weight when viewed in the context of everyday relationship experiences. Being able to keep up with the shifting meanings of social cues and the corresponding needs of their human counterparts, is likely to play a critical role in allowing relational agents to establish and maintain meaningful long-term relationships that develop from relatively superficial encounters to deeper bonds of trust.

Third, our study exclusively relied on language models and conversation transcripts, which omit crucial information such as body language and paralinguistic cues. As generative AI continues expanding in the domains of computer vision and multimodal interfaces (e.g., OpenAI's GPT-4o model, which processes both spoken language and images), future research should test whether artificial agents can reach – or even surpass – the level of accuracy obtained by the human raters with access to the speed dating videos. In addition to boosting LLM's ability to act as relational agents, this technological advancement will also allow for a deeper investigation of the interplay between verbal and nonverbal signals of romantic attraction. While our study highlights the importance of verbal behavior, it is well-established that social impressions are often shaped by an interplay between multiple behavioral channels (Back, 2021; Back & Nestler, 2016; Breil et al., 2021). As LLMs develop a capacity to interpret social cues at different levels, they could be used to explore how the integration of different channels improves predictive accuracy and investigate the impact of mismatched verbal and nonverbal signals (e.g, the impact of enthusiastic language paired with disengaged body language). Investigating such discrepancies may help explain why some interactions appear promising on a verbal level but, in reality, do not lead to genuine romantic interest.

**Conclusion**

Taken together, our work suggests that LLMs like ChatGPT have the capacity to interpret social dynamics in natural conversations, supporting their potential to act as relational agents. As we have noted above, we consider our findings a conservative estimate of AI's capacity to understand human interactions and predict that its ability to interpret conversational dynamics on a much more holistic level is poised to increase significantly over the coming years.

**Declarations:**

- No funding was received for this study.
- The authors declare no conflict of interest.
- The project has received ethical approval from the Ethics Committee of Northwestern University (Protocol #1343-019)
- The manuscript was prepared without the assistance of AI tools.
- Data and analyses codes are available on [OSF](OSF)

# Supplementary Information for "Can Large Language Models Detect Verbal Indicators of Romantic Attraction?"

**Table of Contents**

## Contents

## Supplement Material A: Recruitment flyer

Figure S1. Recruitment Flyer

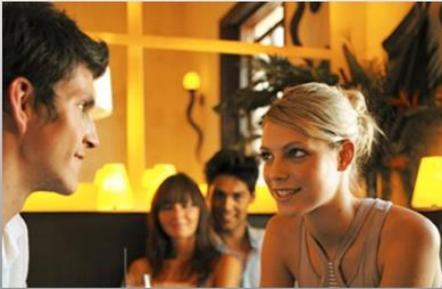

# Supplemental Material B: Measures across participants and observers

### a) Speed Dating Participants

- I am likely to say yes to my interaction partner
- I really liked my interaction partner
- My interaction partner and I seemed to have a lot in common
- My interaction partner and I seemed to have similar personalities
- My interaction partner and I had a real connection

Response format: 9-point scale ranging from 1 = Strongly Disagree to 9 = Strongly Agree.
Level of assessment: Responses were collected for each participant.

### b) ChatGPT 4.0 and Claude 3

LLM Prompt:

*"Below is a transcript of a conversation between two people talking to each other on a speed date. It's the first time they meet.*

- *On a scale from 1-9, how likely do you think the interaction partners will say "yes" to each other and exchange contact information? Only report the numeric value.*
- *On a scale from 1-9, how much do the interaction partners really like each other, with 1 = not at all and 9 = very much. Only report the numeric value.*
- *On a scale from 1-9 how much did the interaction partners have in common, with 1 = nothing at all and 9 = very much? Only report the numeric value.*
- *On a scale from 1-9 how similar were the interaction partners in personality, with 1 = very different and 9 = very similar? Only report the numeric value.*
- *On a scale from 1-9 how much of a real connection did the interaction partners feel, with 1 = no real connection at all and 9 = very strong real connection? Only report the numeric value."*

Level of assessment: Responses were collected for each dyad (i.e., each speed date)

### c) Human judges with access to videos

- He [she] is likely to say "yes" to her [him].
- He [she] really liked her [him].

Response format: 9-point scale ranging from 1 = Strongly Disagree to 9 = Strongly Agree.
Level of assessment: Responses were collected for each participant.

- The interaction partners seemed to have a lot in common
- The interaction partners seemed to have similar personalities
- The interaction partners had a real connection

Response format: 9-point scale ranging from 1 = Strongly Disagree to 9 = Strongly Agree.

Level of assessment: Responses were collected for each dyad (i.e., each speed date)

### d) Human judges with access to transcripts only

- On a scale from 1-9, how likely do you think the interaction partners will say "yes" to each other and exchange contact information?
- On a scale from 1-9, how much do the interaction partners like each other, with 1 = not at all and 9 = very much.
- On a scale from 1-9 how much did the interaction partners have in common, with 1 = nothing at all and 9 = very much?
- On a scale from 1-9 how similar were the interaction partners in personality, with 1 = very different and 9 = very similar?
- On a scale from 1-9 how much of a real connection did the interaction partners feel, with 1 = no real connection at all and 9 = very strong real connection?

Level of assessment: Responses were collected for each dyad (i.e., each speed date)

# Supplemental Material C: Distributions of Ratings

### a) Speed Dating Participants

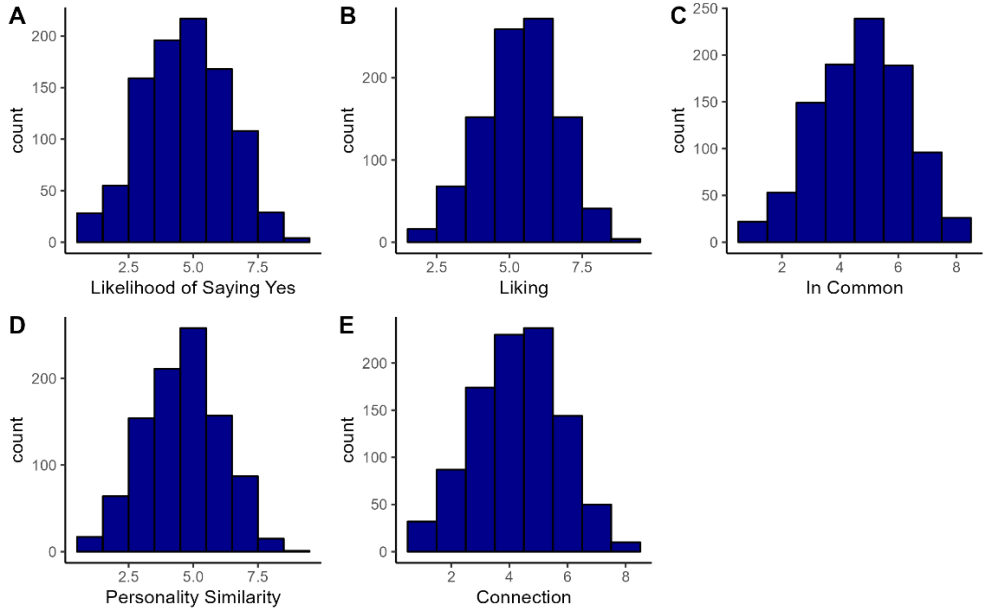

### b) ChatGPT 4.0

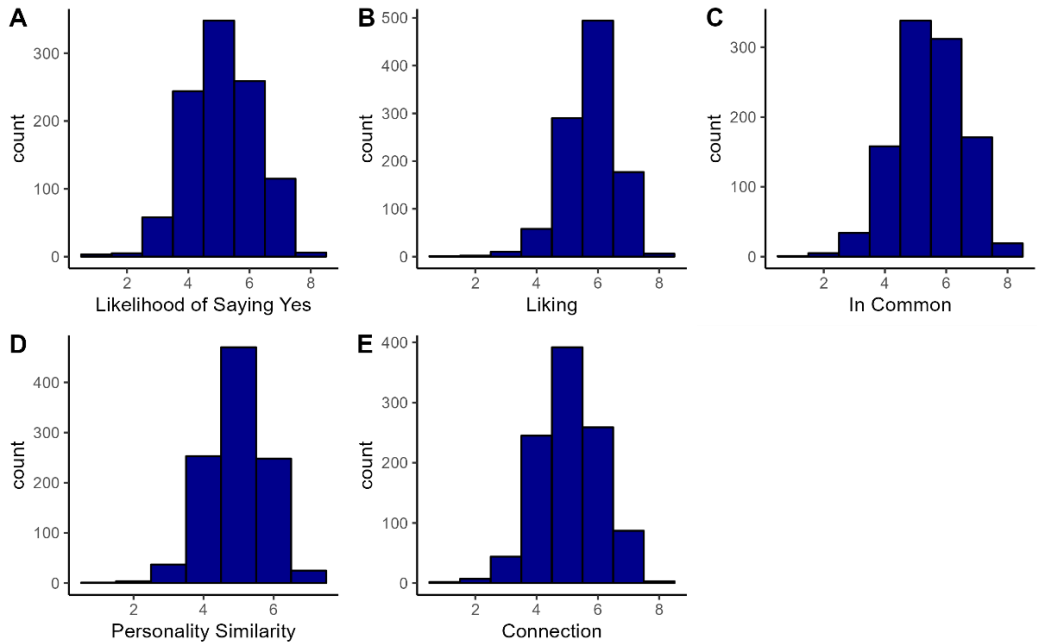

*c) Claude 3*

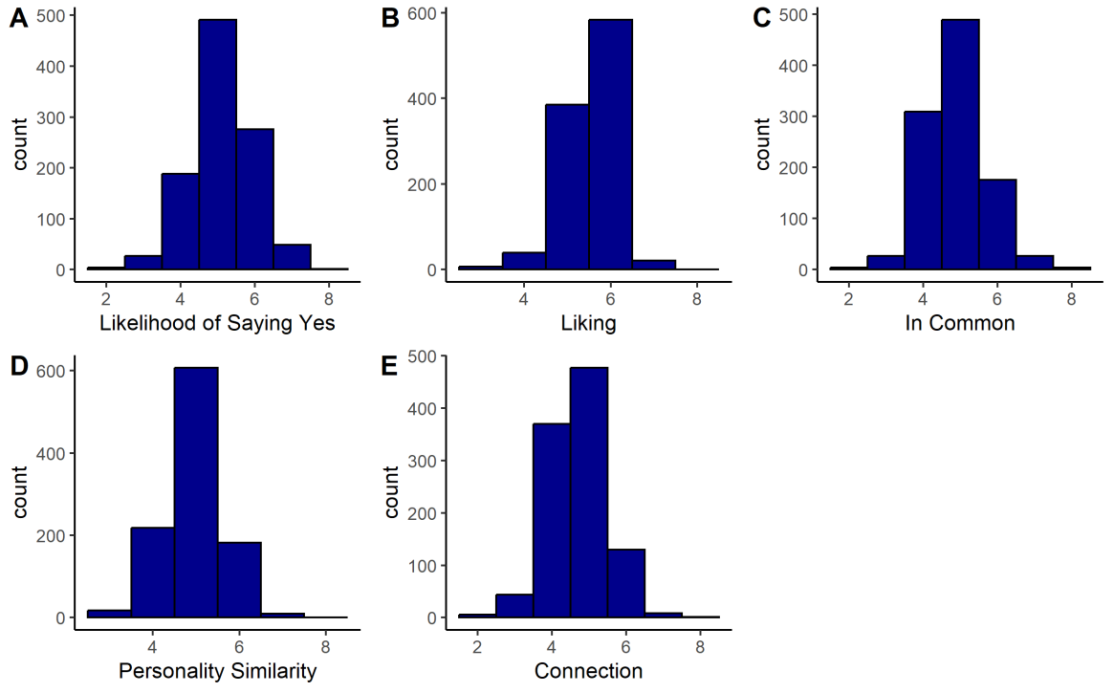

*d) Human judges with access to videos*

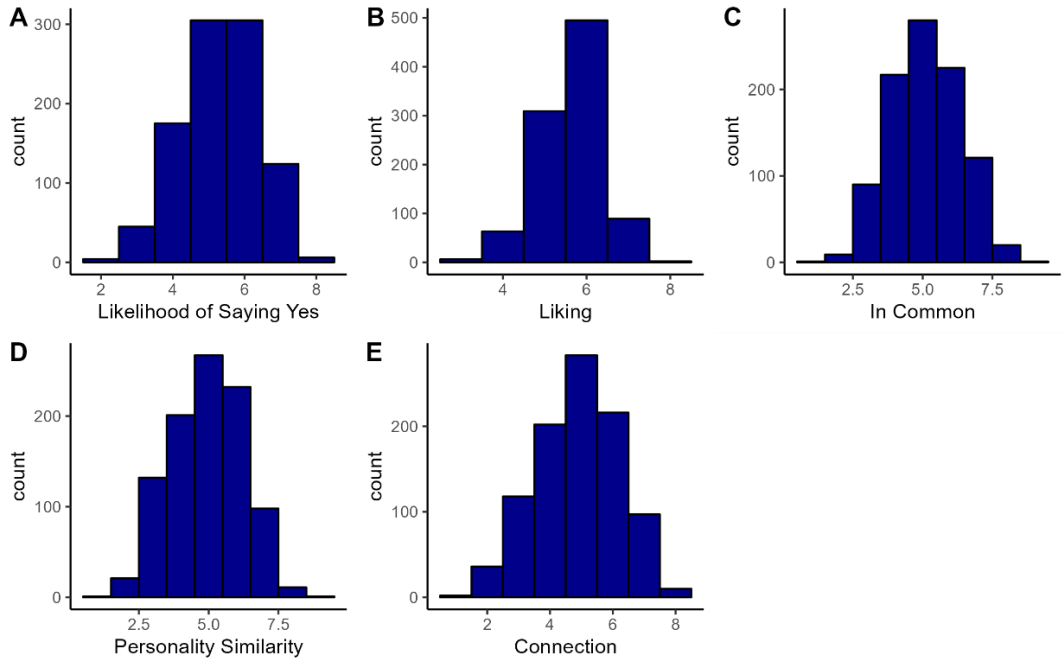

## *e) Human judges with access to transcripts only*

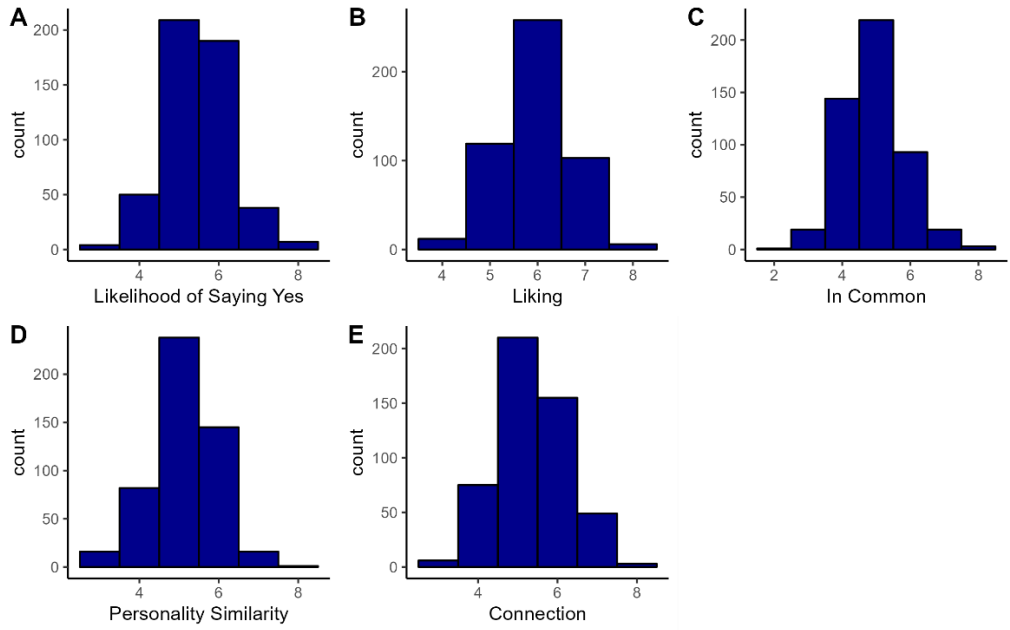

# Supplemental Material D: Intra-class coefficients for all ratings and observers

Table S1. Intra-class coefficients for ratings.

|  | ChatGPT 4.0 | Claude 3 | Human judges (video) | Human judges (transcripts) |
|---|---|---|---|---|
| Likelihood of saying Yes | ICC = 0.80 | ICC = 0.73 | ICC = 0.92 | ICC = 0.49 |
| Liking | ICC = 0.76 | ICC = 0.71 | ICC = 0.89 | ICC = 0.46 |
| In common | ICC = 0.75 | ICC = 0.79 | ICC = 0.94 | ICC = 0.51 |
| Personality similarity | ICC = 0.65 | ICC = 0.64 | ICC = 0.93 | ICC = 0.45 |
| Connection | ICC = 0.75 | ICC = 0.68 | ICC = 0.94 | ICC = 0.40 |

# Supplemental Material E: Pearson zero-order correlation of all variables

Table S2. Zero-order correlations.

|  | 1 | 2 | 3 | 4 | 5 | 6 | 7 | 8 | 9 | 10 | 11 | 12 | 13 | 14 | 15 | 16 | 17 | 18 | 19 | 20 | 21 | 22 | 23 | 24 | 25 |
|---|---|---|---|---|---|---|---|---|---|---|---|---|---|---|---|---|---|---|---|---|---|---|---|---|---|
| 1. Matching (SR) | - | | | | | | | | | | | | | | | | | | | | | | | | |
| 2. Liking (SR) | .75 | - | | | | | | | | | | | | | | | | | | | | | | | |
| 3. Exchange (SR) | .50 | .41 | - | | | | | | | | | | | | | | | | | | | | | | |
| 4. Common (SR) | .59 | .64 | .33 | - | | | | | | | | | | | | | | | | | | | | | |
| 5. Similarity (SR) | .64 | .70 | .39 | .79 | - | | | | | | | | | | | | | | | | | | | | |
| 6. Connection (SR) | .67 | .71 | .43 | .71 | .80 | - | | | | | | | | | | | | | | | | | | | |
| 7. Matching (GPT) | .13 | .15 | .12 | .23 | .19 | .17 | - | | | | | | | | | | | | | | | | | | |
| 8. Liking (GPT) | .12 | .13 | .11 | .19 | .15 | .13 | .90 | - | | | | | | | | | | | | | | | | | |
| 9. Common (GPT) | .09 | .12 | .09 | .23 | .14 | .11 | .76 | .79 | - | | | | | | | | | | | | | | | | |
| 10. Similarity (GPT) | .12 | .12 | .10 | .21 | .16 | .14 | .78 | .81 | .82 | - | | | | | | | | | | | | | | | |
| 11. Connection (GPT) | .13 | .15 | .10 | .23 | .18 | .16 | .91 | .91 | .84 | .85 | - | | | | | | | | | | | | | | |
| 12. Matching (Claude) | .12 | .15 | .07 | .2 | .18 | .16 | .4 | .37 | .27 | .33 | .38 | - | | | | | | | | | | | | | |
| 13. Liking (Claude) | .04 | .09 | .05 | .13 | .12 | .11 | .39 | .38 | .28 | .32 | .37 | .71 | - | | | | | | | | | | | | |
| 14. Common (Claude) | .11 | .16 | .1 | .29 | .21 | .2 | .47 | .43 | .41 | .42 | .45 | .61 | .67 | - | | | | | | | | | | | |
| 15. Similarity (Claude) | .08 | .12 | .08 | .23 | .17 | .17 | .43 | .41 | .37 | .39 | .42 | .57 | .76 | .84 | - | | | | | | | | | | |
| 16. Connection (Claude) | .08 | .13 | .05 | .22 | .19 | .16 | .45 | .42 | .33 | .36 | .42 | .74 | .85 | .81 | .82 | - | | | | | | | | | |
| 17. Matching (Coder V) | .40 | .44 | .31 | .43 | .44 | .47 | .31 | .29 | .24 | .26 | .30 | .22 | .18 | .23 | .19 | .21 | - | | | | | | | | |
| 18. Liking (Coder V) | .31 | .39 | .28 | .42 | .40 | .44 | .24 | .21 | .17 | .19 | .22 | .2 | .14 | .22 | .18 | .19 | .82 | - | | | | | | | |
| 19. Connection (Coder V) | .35 | .45 | .32 | .48 | .49 | .49 | .27 | .24 | .21 | .23 | .27 | .24 | .19 | .29 | .23 | .25 | .88 | .79 | - | | | | | | |
| 20. Common (Coder V) | .23 | .31 | .23 | .55 | .42 | .38 | .28 | .24 | .32 | .29 | .30 | .27 | .21 | .41 | .33 | .29 | .58 | .54 | .72 | - | | | | | |
| 21. Similarity (Coder V) | .31 | .39 | .28 | .44 | .46 | .43 | .23 | .2 | .19 | .21 | .23 | .21 | .15 | .25 | .2 | .2 | .77 | .68 | .88 | .71 | - | | | | |
| 22. Matching (Coder T) | .27 | .29 | .13 | .40 | .36 | .34 | .34 | .35 | .25 | .29 | .34 | .29 | .28 | .36 | .32 | .34 | .45 | .41 | .45 | .42 | .37 | - | | | |
| 23. Liking (Coder T) | .23 | .26 | .15 | .35 | .32 | .30 | .33 | .33 | .23 | .24 | .31 | .24 | .25 | .3 | .27 | .3 | .41 | .39 | .43 | .37 | .36 | .89 | - | | |
| 24. Connection (Coder T) | .23 | .26 | .13 | .40 | .34 | .32 | .27 | .29 | .22 | .23 | .28 | .23 | .22 | .35 | .28 | .31 | .41 | .41 | .46 | .45 | .35 | .85 | .80 | - | |
| 25. Common (Coder T) | .16 | .17 | .11 | .45 | .29 | .24 | .29 | .29 | .36 | .35 | .32 | .26 | .21 | .45 | .36 | .32 | .25 | .29 | .32 | .61 | .26 | .58 | .52 | .63 | - |
| 26. Similarity (Coder T) | .24 | .26 | .15 | .42 | .37 | .33 | .31 | .32 | .27 | .27 | .31 | .25 | .23 | .34 | .3 | .31 | .41 | .40 | .47 | .49 | .42 | .80 | .77 | .81 | .64 |

SR = Participants' self-reports, GPT = Predictions made by ChatGPT 4, Claude = Predictions made by Claude 3, Coder V = Coders with access to video recordings, Coder T = Coders with access to text transcripts only.

# Supplemental Material F: Findings for Claude 3

Fig. S2 Correlations between the speed-dating ratings of the two LLMs and different evaluators

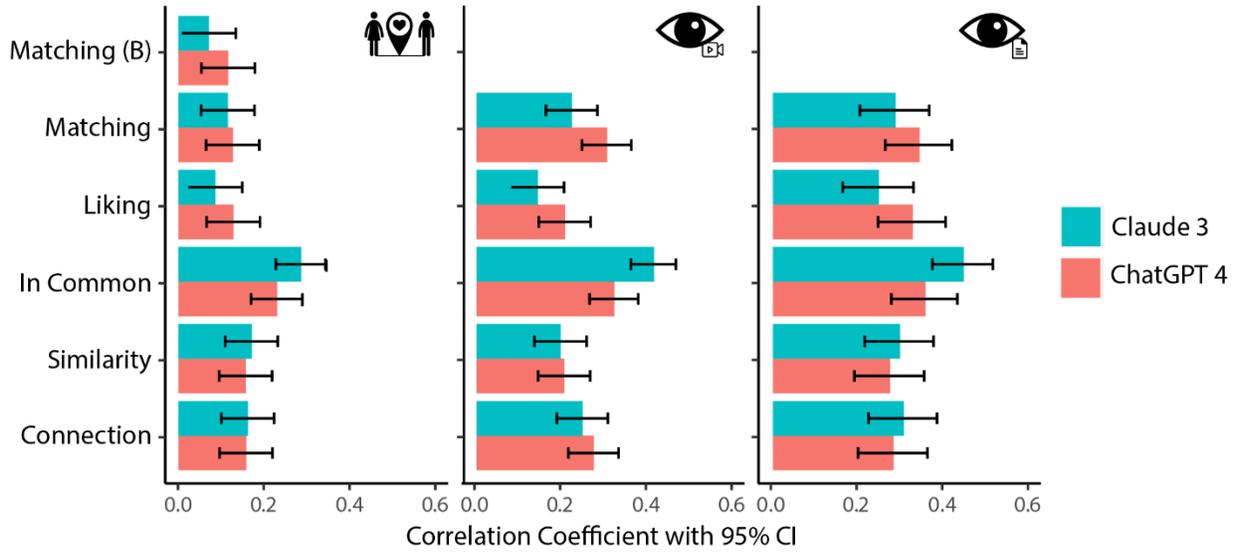

## Supplemental Material G: LIWC dimensions (True Negatives)

LIWC dimensions that were neither utilized by ChatGPT nor actually valid indicators of romantic attraction (True Negatives).

| | |
|---|---|
| Culture | Acquire |
| Anger | Negative emotions |
| Affect | Attention |
| Health | Adjectives |
| 1st person plural | Assent |
| Total Pronouns | Determiners |
| Space (Perception) | 2nd person |
| Communication | Words per sentence |
| Prosocial behavior | Lack |
| Articles | Illness |
| Male references | Need |
| Discrepancy | Reward |
| Leisure | Fatigue |
| Mental health | Tentative |
| Wellness | Filler |
| Prepositions | Visual |
| Periods | Curiosity |
| Drives (Psychological processes) | Politeness |
| Religion | Impersonal pronouns |
| Netspeak | Conflict |
| Negative Tone | Food |
| Friends | Physical |
| All Punctuation | Quantity |
| Perception | Moralization |
| Big Words (7+ letters) | Achievement |
| Other punctuation | Comma |
| Power | Family |
| Adverbs | Conjunctions |
| Anxiety | Numbers |
| Fulfilled | 1st person |
| Political | Time |
| Memory | Risk |
| Authentic | Dictionary words |
| Nonfluencies | Sexual |
| Clout | |
| Auditory | |
| Death | |
| Linguistic dimensions | |
| Ethnicity | |

## Supplemental Material H: ChatGPT prompt for taxonomy development

Prompt for ChatGPT to create a parsimonious taxonomy of mutual attraction indicators from the first 100 explanations. The number of explanations was limited to 100 due to the limited token input available for GPT4o:

*"Below are 100 explanations for why a speed date was successful or not. Create a parsimonious taxonomy of explanation categories that captures recurring explanations and are psychologically meaningful:*

*[Explanations piped in]"*

The table below shows the taxonomy suggested by ChatGPT in response to the above prompt. We manually removed the indicators in red (i.e., "External Factors and Contextual Influences", "Internal States and Perceptions", and "Non-Verbal and Unspoken Cues") as they are difficult to reliably classify in transcripts.

# Supplemental Material I: Taxonomy capturing common indicators of mutual attraction as created by ChatGPT (and updated manually)

Table S3. Taxonomy of romantic attraction.

| Main category | Indicator | Description |
| --- | --- | --- |
| Conversation Flow and Engagement | Smoothness of Conversation | Whether the conversation flows naturally without awkward pauses or forced topics. |
| Conversation Flow and Engagement | Mutual Engagement | Degree to which both participants are actively involved, asking questions, and showing interest. |
| Conversation Flow and Engagement | Humor and Playfulness | Presence of shared jokes, teasing, or playful banter that indicates comfort and enjoyment. |
| Shared Interests and Values | Common Interests | Shared hobbies, academic pursuits, or extracurricular activities that create a connection. |
| Shared Interests and Values | Aligned Values and Goals | Similar future aspirations, lifestyle preferences, or personal values that suggest compatibility. |
| Emotional and Personal Connection | Emotional Depth | Conversations that go beyond surface-level topics to include personal stories, emotions, or vulnerabilities. |
| Emotional and Personal Connection | Reciprocity of Disclosure | Balanced sharing of personal information, indicating mutual trust and interest. |
| Emotional and Personal Connection | Flirtation and Romantic Signals | Presence of flirting, compliments, or subtle romantic cues that suggest potential attraction. |
| Perceived Compatibility and Future Potential | Perceived Compatibility | Participants' subjective feelings of how well they "click" based on conversation dynamics. |
| Perceived Compatibility and Future Potential | Expression of Future Intentions: | Discussions about meeting again, exchanging contact information, or planning future activities together. |
| Perceived Compatibility and Future Potential | Social Compatibility | Mutual acquaintances, similar social circles, or shared experiences that enhance the likelihood of a future relationship. |
| External Factors and Contextual Influences | Event Context and Pressure | Influence of the speed dating context itself, including the time constraint and the setting, on conversation outcomes. |
| External Factors and Contextual Influences | Influence of Third Parties | Mention of mutual friends or shared social environments that might sway the decision to exchange contact information. |
| Internal States and Perceptions | Self-Perception and Confidence | How each participant perceives themselves and their confidence level during the conversation. |
| Internal States and Perceptions | Perception of the Other | Initial impressions and ongoing assessments of the other person's attractiveness, personality, and potential as a partner. |
| Conversational Challenges | Moments of Tension or Disagreement | Instances of conflict, misunderstandings, or awkwardness that might derail the conversation. |
| Conversational Challenges | Mismatch in Interests or Values | Divergences in interests, life goals, or values that could indicate long-term incompatibility. |
| Non-Verbal and Unspoken Cues | Tone and Body Language | Although not explicitly detailed in the text, implied tone and the likely impact of body language and non-verbal cues. |
| Non-Verbal and Unspoken Cues | Subtle Cues and Ambiguity | Ambiguous responses or lack of clarity in communication that might affect perceptions. |

# Supplemental Material J: ChatGPT prompt for explanation classification

Prompt for ChatGPT to classify each of its explanations for speed dates according to the taxonomy:

*"You predicted the success of a speed date based on a transcript of the conversation partners. Does the following explanation you gave for your prediction reference one or more of the following indicators:*

- *Smoothness of conversation: The extent to which the conversation flows naturally without awkward pauses or forced topics.*
- *Mutual Engagement: The degree to which both participants are actively involved, asking questions, and showing interest.*
- *Humor and Playfulness: The presence of shared jokes, teasing, or playful banter that indicates comfort and enjoyment.*
- *Common Interests: Shared hobbies, academic pursuits, or extracurricular activities that create a connection.*
- *Aligned Values and Goals: Similar future aspirations, lifestyle preferences, or personal values that suggest compatibility.*
- *Emotional Depth: Conversations that go beyond surface-level topics to include personal stories, emotions, or vulnerabilities.*
- *Reciprocity of Disclosure: Balanced sharing of personal information, indicating mutual trust and interest.*
- *Flirtation and Romantic Signals: Presence of flirting, compliments, or subtle romantic cues that suggest potential attraction.*
- *Perceived Compatibility: Participants' subjective feelings of how well they click based on conversation dynamics.*
- *Expression of Future Intentions: Discussions about meeting again, exchanging contact information, or planning future activities together.*
- *Social Compatibility: Mutual acquaintances, similar social circles, or shared experiences that enhance the likelihood of a future relationship.*
- *Moments of Tension or Disagreement: Instances of conflict, misunderstandings, or awkwardness that might derail the conversation.*
- *Mismatch in Interests or Values: Divergences in interests, life goals, or values that could indicate long-term incompatibility.*

*Please assign a 1 if the explanation mentions the construct in question and a 0 if it does not. Use the following structure for your output*

*Indicator name:*
*Indicator number:*
*Is the indicator present or not with 1 = yes and 0 = no:*
*Here is the explanation: [Explanation dynamically piped in]"*

# Supplemental Material K: ChatGPT prompt for taxonomy rating.

Prompt for ChatGPT rating of taxonomy. The questions were generated by ChatGPT based on the description of the taxonomy indicators.

*"Below is a transcript of a conversation between two people talking to each other on a speed date. It's the first time they meet.*

- *On a scale from 1-9, how likely do you think the interaction partners will say "yes" to each other and exchange contact information? Only report the numeric value.*
- *On a scale from 1-9, how smoothly did the conversation progress without awkward pauses or forced topics? Only report the numeric value.*
- *On a scale from 1-9, to what extent did both conversation partners actively participate and show interest in the conversation? Only report the numeric value.*
- *On a scale from 1-9, how much did humor or playful banter contribute to a sense of comfort and enjoyment during the conversation? Only report the numeric value.*
- *On a scale from 1-9, how much common ground did the conversation partners discover in terms of shared hobbies, activities, or interests? Only report the numeric value.*
- *On a scale from 1-9, how aligned did the conversation partners' values and future goals seem? Only report the numeric value.*
- *On a scale from 1-9, to what extent did the conversation go beyond surface-level topics and delve into more personal or emotional areas? Only report the numeric value.*
- *On a scale from 1-9, how balanced was the sharing of personal information and stories between the conversation partners? Only report the numeric value.*
- *On a scale from 1-9, how evident were flirtatious cues or romantic signals in the conversation? Only report the numeric value.*
- *On a scale from 1-9, how compatible did the conversation partners seem based on their conversation? Only report the numeric value.*
- *On a scale from 1-9, did the conversation partners express interest in meeting again or continuing the conversation outside of the speed date? Only report the numeric value.*
- *On a scale from 1-9, how well did the conversation partners' social circles or experiences seem to overlap, potentially making future interactions more likely? Only report the numeric value.*
- *On a scale from 1-9, were there any moments of tension, disagreement, or awkwardness that negatively affected the interaction? Only report the numeric value.*
- *On a scale from 1-9, were there any significant differences in interests or values that might hinder a future connection between the two conversation partners? Only report the numeric value."*